\documentclass[letterpaper, 10 pt, journal, twoside]{IEEEtran}
\IEEEoverridecommandlockouts

\usepackage{fancyhdr}
\usepackage[space]{cite}
\usepackage{amsmath,amsopn,amstext,amsfonts,mathtools}
\usepackage{amssymb}
\usepackage{float}
\usepackage{siunitx}
\usepackage{graphicx}
\usepackage{textcomp}
\usepackage{xcolor}
\usepackage{multirow}
\usepackage{booktabs}
\usepackage{svg}
\usepackage[tight]{subfigure}
\usepackage{epstopdf}
\usepackage[linesnumbered,boxed,ruled,commentsnumbered]{algorithm2e}
\usepackage{algpseudocode}
\usepackage[font=small,skip=5pt]{caption}
\usepackage{threeparttable}
\usepackage{geometry}
\def\BibTeX{{\rm B\kern-.05em{\sc i\kern-.025em b}\kern-.08em
  T\kern-.1667em\lower.7ex\hbox{E}\kern-.125emX}}

\begin{document}

\title{Large-Scale LiDAR Consistent Mapping using Hierarchical LiDAR Bundle Adjustment}

\author{Xiyuan Liu, Zheng Liu, Fanze Kong, and Fu Zhang
\thanks{Xiyuan Liu, Zheng Liu, Fanze Kong and Fu Zhang are with the Department of Mechanical Engineering, The University of Hong Kong, Hong Kong. (email: {\tt\footnotesize $\{$xliuaa,u3007335,kongfz$\}$@connect.hku.hk}, {\tt\footnotesize $ $fuzhang$ $@hku.hk}) (\textit{Corresponding author: Fu Zhang}).}}
\maketitle
\thispagestyle{empty}

\begin{abstract}
Reconstructing an accurate and consistent large-scale LiDAR point cloud map is crucial for robotics applications. The existing solution, pose graph optimization, though it is time-efficient, does not directly optimize the mapping consistency. LiDAR bundle adjustment (BA) has been recently proposed to resolve this issue; however, it is too time-consuming on large-scale maps. To mitigate this problem, this paper presents a globally consistent and efficient mapping method suitable for large-scale maps. Our proposed work consists of a bottom-up hierarchical BA and a top-down pose graph optimization, which combines the advantages of both methods. With the hierarchical design, we solve multiple BA problems with a much smaller Hessian matrix size than the original BA; with the pose graph optimization, we smoothly and efficiently update the LiDAR poses. The effectiveness and robustness of our proposed approach have been validated on multiple spatially and timely large-scale public spinning LiDAR datasets, i.e., KITTI, MulRan and Newer College, and self-collected solid-state LiDAR datasets under structured and unstructured scenes. With proper setups, we demonstrate our work could generate a globally consistent map with around 12$\%$ of the sequence time.
\end{abstract}


\section{Introduction}

Reconstructing the three-dimensional (3D) high-resolution map is of great significance in the fields of robotics, environmental and civil engineering. This 3D map could be used as a prior for autonomous service robots and as an information model for buildings and geographical measurements. Compared with the traditional 3D laser scanner, the light detection and ranging (LiDAR) sensor extraordinarily fits into this purpose due to its fast scanning rate. Moreover, it is more lightweight, cost-effective and flexible to be carried on multiple platforms, e.g., ground or aerial vehicles and hand-held devices. In this paper, we focus on developing an accurate and consistent LiDAR mapping method for large-scale maps.

Rich research results have been presented on LiDAR-based mapping algorithms~\cite{fast_lio2,lio_sam,mulls}, which generate both point cloud maps and LiDAR odometry. Due to the accumulation of scan-to-map registration errors, odometry drift usually appears and further leads to divergence in the point cloud map. The most well-known method to refine the mapping quality (closing the gap) is pose graph optimization, which minimizes the relative pose errors between two LiDAR frames. In pose graph optimization, the relative pose estimation is assumed to follow Gaussian distribution. However, this approximation might be overestimated in reality~\cite{gpu_accelerated}. Moreover, the pose graph optimization does not directly optimize the consistency of the point cloud that the divergence within the point cloud map might only be narrowed but not fully eliminated (or not even aware of). This phenomenon is more obvious when the wrong loops are detected, or incorrect relative transformation estimations happen.

LiDAR bundle adjustment (BA) approach~\cite{balm,zhou2021plane} directly optimizes the mapping consistency by minimizing the overall point-to-plane distance, which often leads to high mapping quality that is necessary for mapping applications. In~\cite{balm}, the plane parameters are analytically solved first such that the final optimization problem is only related to the LiDAR pose. In~\cite{zhou2021plane}, the plane parameters are eliminated in each iteration of the optimization by a Schur complement trick as in visual BA~\cite{zhang2012graph,tian2021hier}. Either way, the resultant optimization is (at least) the dimension of the LiDAR pose number $N$, requiring $O(N^3)$ time to solve~\cite{newba}. The cubic growth of the computation time has prohibited the bundle adjustment for large-scale maps with large pose numbers.

\begin{figure*}[ht]
  \centering
  \includegraphics[width=1.0\linewidth]{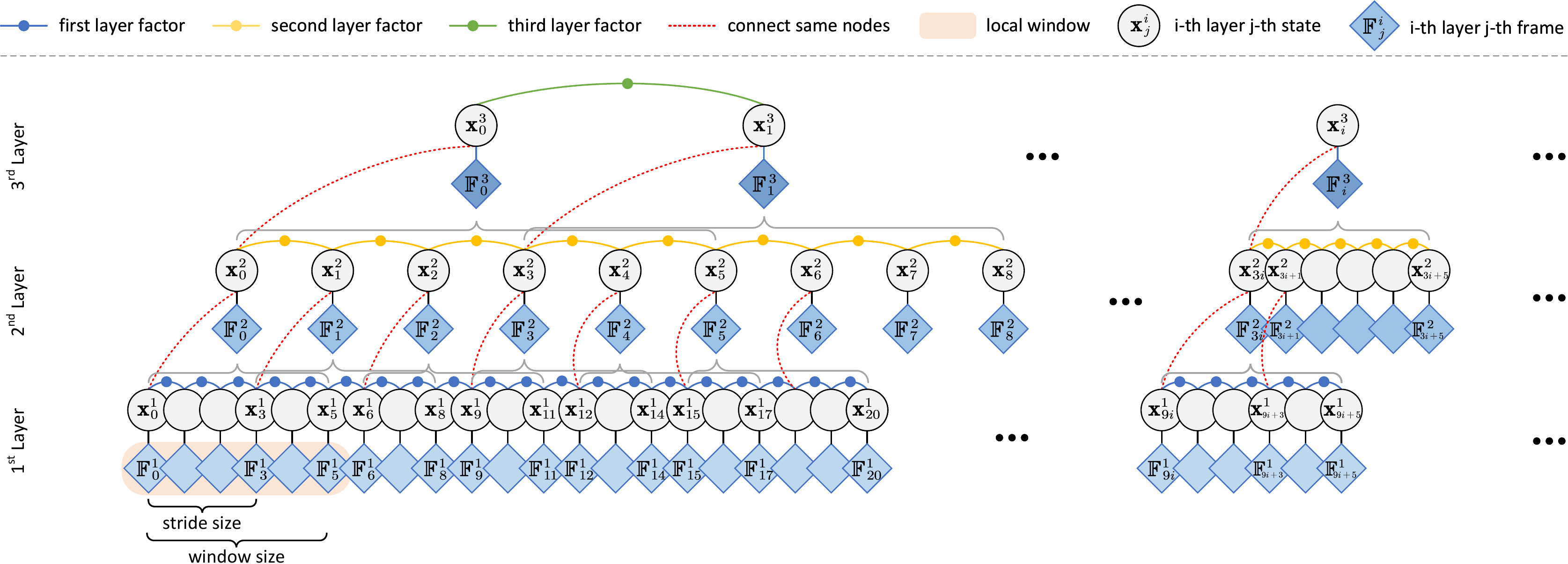}
  \caption{Pyramid structure of the proposed hierarchical bundle adjustment with \textit{layer number l=3}, \textit{stride size s=3} and \textit{window size w=6}. The first, second and third layer factors connect every two adjacent nodes within the same layer. The red dashed line connects the nodes ought to be the same, e.g., $\mathbf{x}^1_{9i}$, the first node of the local window from the lower layer to the node constructed from this local window from the upper layer, e.g., $\mathbf{x}^2_{3i}$ and $\mathbf{x}^3_i$.}
  \vspace{-0.4cm}
  \label{fig:lidar_frame}
\end{figure*}

To address the above issues, we propose a hierarchical LiDAR BA method to globally optimize the mapping consistency while maintaining time efficiency. The method constructs a pyramid structure of frame poses (see Fig.~\ref{fig:lidar_frame}) and conduct a bottom-up hierarchical bundle adjustment and a top-down pose graph optimization (see Fig.~\ref{fig:overview}). The bottom-up process conducts a hierarchical bundle adjustment within local windows from the bottom layer (local BA) to the top layer frames (global BA). Such design benefits the computation time since the process of local BA in each layer is suitable for parallel processing and the time complexity of each local BA is relatively low due to the small number of poses involved. One issue in the bottom-up process is that it neglects features co-visible across different local windows, which could lower the accuracy. To mitigate this issue, the top-down process constructs a pose graph from the top to the bottom layers and distributes the errors by pose graph optimization. The two process iterates until convergence.

With the hierarchical bundle adjustment design, we could both directly optimize the consistency of the planar surfaces within the point cloud and avoid solving a cost function with large dimensions. With pose graph optimization, we properly update the entire LiDAR poses towards convergence in a fast and reliable manner. To retain the smoothness between every two adjacent keyframes, we keep an overlap area between them by setting the stride size smaller than the window size. To further boost the optimization speed, we have applied a filter to remove outlier points and implemented CPU-based parallel processing when constructing the pyramid. In summary, our contributions are as follows:

\begin{itemize}
  \item We propose a hierarchical bundle adjustment method to globally optimize the LiDAR mapping consistency and odometry accuracy. Our proposed approach improves the mapping quality given a good initial pose trajectory (e.g., from a pose graph optimization) and even closes the gap when the initial pose trajectory has large drifts.
  \item The effectiveness of our proposed work has been validated on multiple public mechanical spinning LiDAR datasets and our self-collected solid-state LiDAR dataset in both structured and unstructured scenes.
\end{itemize}

\vspace{-0.2cm}
\section{Related Works}

Multiple approaches have been discussed in the literature on improving the mapping quality, which could be mainly divided into two categories: pose graph optimization and plane (bundle) adjustment-based methods. In pose graph optimization, the relative transformation (pose constraint) between two frames is estimated by ICP~\cite{ICP} or its variants~\cite{GICP,TASER}. This relative pose error is then weighted by the information matrix, which is usually the inverse of the corresponding Hessian~\cite{hess2016loop} or simply a constant matrix~\cite{behley2018rss}. The pose graph is optimized when the summed relative pose errors are minimized. Though computationally efficient, one important issue of pose graph optimization is that it does not directly optimize the consistency of the point cloud. Due to incorrect estimation or imprecise modelling of the relative pose constraint, the pose graph might converge to a local minimum that
considerable divergence within the point cloud could still exist~\cite{mulls,liang2021novel}.

The plane adjustment method directly optimizes the consistency of the point cloud by minimizing the summed point-to-plane distance. In plane adjustment, each plane feature is represented by two parameters, i.e., the distance from its center to the viewpoint and the estimated plane normal vector~\cite{zhou2021plane}. In~\cite{large_scale}, authors concurrently optimize both the LiDAR poses and the geometric plane features. This method needs to maintain and update the parameters of all features during the optimization, whereas the total number of features will rapidly grow when the scale of the map enlarges, leaving a huge dimension of the cost function to solve. {Though with the Schur complement technique, the optimization variables could be reduced to LiDAR poses only, this method is prone to generate glitches in pose estimation during optimization in real-world practices.}

The bundle adjustment method improves the technique of plane adjustment by eliminating the feature parameter prior to the optimization using a closed-form solution~\cite{balm}. In~\cite{balm}, authors segment the point cloud into multiple voxels, each containing a plane feature. The original point-to-plane minimization problem is transformed into the minimization of the eigenvalue of points covariance in each voxel. Such a method needs to iterate through every point within each feature to derive the Hessian matrix, whose time complexity is the square of the number of points, causing a great computation demand. This problem is addressed in the following-up work~\cite{newba}, which aggregates all the points of a feature observed by the same pose and fundamentally removes the dependence of time complexity on the point number. {Despite this, in all the above-mentioned plane and bundle adjustment methods, the computation complexity is still cubic to the pose number, which is not practical for large-scale maps.} Moreover, when the divergence in the map is larger than or approximates the maximum voxel size, these methods might have a slow convergence rate.

Our proposed hierarchical bundle adjustment approach takes advantage of both BA and pose graph optimization. We use BA to minimize the point-to-plane distance directly and pose graph optimization to smoothly and efficiently update the LiDAR poses to avoid glitches in pose estimation. With the hierarchical design, we could parallelly solve multiple BA problems with a much smaller Hessian matrix size compared with the original problem using~\cite{balm}. Moreover, we could flexibly set BA parameters from the bottom layers to the top layers in accordance with the quality of the initial pose trajectory.

\vspace{-0.5cm}
\section{Methodology}

\subsection{Overview}

\begin{figure}[ht]
  \centering
  \includegraphics[width=1.0\linewidth]{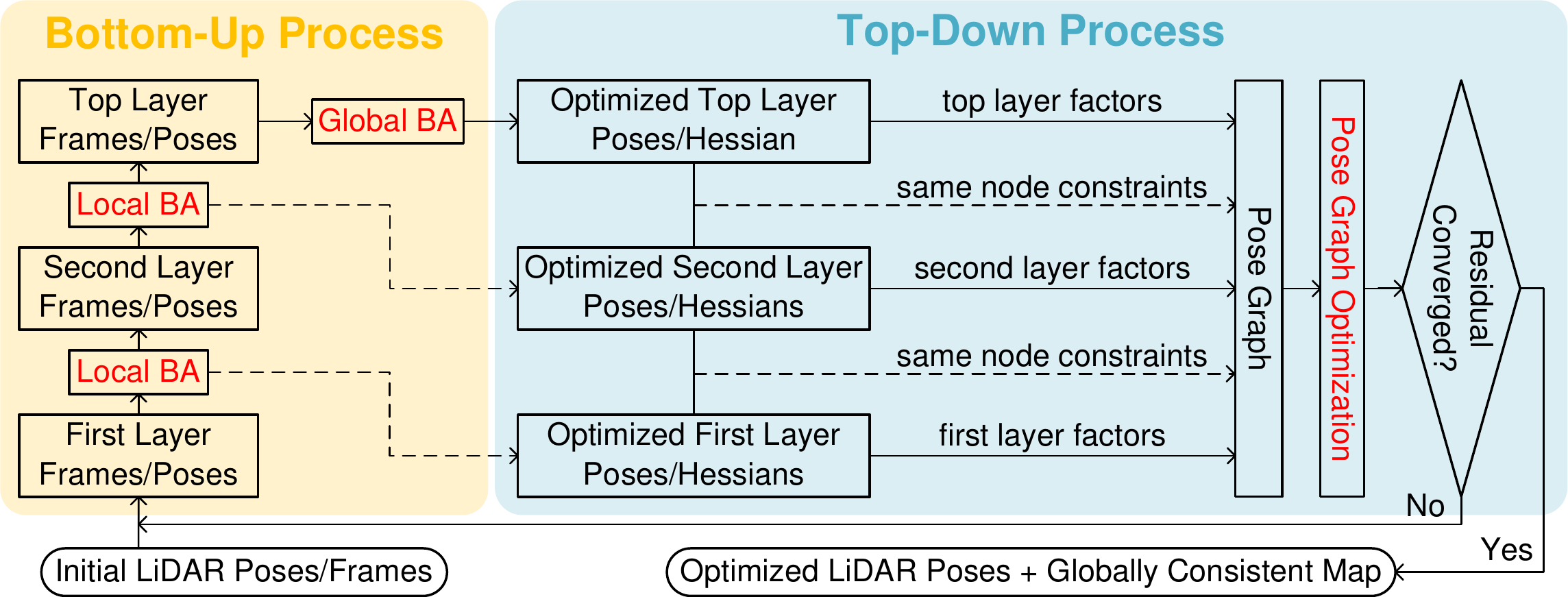}
  \caption{System overview. The light yellow region depicts the bottom-up process and the light blue region depicts the top-down process.}
  \vspace{-0.4cm}
  \label{fig:overview}
\end{figure}

The system workflow of our proposed method is illustrated in Fig.~\ref{fig:overview}. The inputs are raw or deskewed points from each LiDAR scan and an initial estimation of their corresponding poses in the global frame, which could be obtained from the general LiDAR odometry or simultaneous locomotion and mapping (SLAM) algorithms. The method consists of two processes, bottom-up (see Sec.~\ref{sec:bottom_up}) and top-down (see Sec.~\ref{sec:top_down}), which iterate until convergence. In the bottom-up process, a local BA is performed on LiDAR frames within smaller local windows to construct keyframes from the first layer to the second layer (see Fig.~\ref{fig:lidar_frame}). This process is hierarchically performed until the optimal layer number is met and a global BA is performed on the top layer keyframes. Then a pose graph is constructed using the factors contributed from each optimized layer and between adjacent layers (see Fig.~\ref{fig:lidar_frame}).

As shown in Fig.~\ref{fig:lidar_frame}, the term \textit{first layer}, also referred to \textit{bottom layer} in the context above, describes the collection of initial LiDAR frames and poses. Similarly, the term \textit{second layer} represents the collection of LiDAR keyframes and poses created from the first layer using the local BA. The term \textit{top layer} means the collection of the last remaining LiDAR keyframes (in Fig.~\ref{fig:lidar_frame}, the top layer refers to the third layer). The process of hierarchically creating LiDAR keyframes from the bottom layer to the top layer is called the bottom-up process. The process of updating bottom layer LiDAR poses by pose graph optimization is called the top-down process.

\vspace{-0.2cm}
\subsection{Bottom-Up Hierarchical BA}\label{sec:bottom_up}

We denote $\mathbb{F}^i_j$ the $j$-th LiDAR frame of the $i$-th layer and $\mathbf{x}^i_j\triangleq\mathbf{T}^i_j=\left(\mathbf{R,t}\right)\in$ SE(3) its corresponding pose. We denote $\mathbf{T}^i_{j,k}$ the relative pose between $\mathbf{T}^i_j$ and $\mathbf{T}^i_k$, i.e., $\mathbf{T}^i_{j,k}=\left(\mathbf{T}^i_j\right)^{-1}\cdot\mathbf{T}^i_k$. It is noted that points in $\mathbb{F}^i_j$ is represented in the LiDAR local frame, and $\mathbf{T}^i_j$ is in the global frame. We denote $w$ as the local window size and $s$ as the stride size during the bottom-up construction of the LiDAR keyframes, as illustrated in Fig.~\ref{fig:lidar_frame}. Assume we have $N_i$ total number of LiDAR frames from the $i$-th layer. In the bottom-up process, a local BA is performed in each local window using the provided initial pose trajectory, and the relative pose between each frame and the first frame in this window is optimized. The derived Hessian matrix $\mathbf{H}$ from the BA in each local window is also recorded and used as the information matrix in the latter top-down pose graph construction. Given a local window of $w$ LiDAR frames $\{\mathbb{F}^i_{sj+k}\mid j=0,\cdots,\lfloor\frac{N_i-w}{s}\rfloor;k=0,\cdots,w-1\}$ in $i$-th layer and its optimized relative poses $\mathbf{T}^{i\ast}_{j,k}$ in the window, we aggregate these frames into a keyframe for the ($i$+1)-th layer. Points in this keyframe are all transformed into the first frame of the local window, and the pose of the keyframe, denoted by $\mathbf{T}^{i+1}_j$, is set as the pose of the first frame optimized in the preceding local window, i.e.,
\begin{equation}\label{eq:FT}
  \begin{aligned}
  \mathbb{F}^{i+1}_j&\triangleq\bigcup^{w-1}_{k=0}\left(\mathbf{T}^{i\ast}_{s\cdot j,s\cdot j+k}\cdot\mathbb{F}^i_{s\cdot j+k}\right)\\
  \mathbf{T}^{i+1}_j&=\mathbf{T}^{i\ast}_{s\cdot j}=\prod^j_{k=1}\mathbf{T}^{i\ast}_{s\cdot k-s,s\cdot k}
  \end{aligned}
\end{equation}

This process could be repeatedly performed from the lower layer to the upper layer until the optimal layer number $l$ is reached. It is noticed that the construction of new keyframes (local BA) does not rely on the frames outside the local window, making it suitable to use multiple local windows for parallel processing in the same layer. Suppose we have $N$ total number of LiDAR frames, i.e., $N_1=N$, and each time we choose to aggregate $w$ frames from the lower layer into one frame to the upper layer with stride frame size $s$. Let $n$ be the number of threads that could be used for parallel processing. Since the computation time of BA is $O\left(M^3\right)$ with $M$ is the number of involved poses, we could derive the overall time consumption $O\left(T_l\right)$ for a $l$-th layer pyramid.

The total time consumption of $l$-th layer pyramid includes that consumed by the local BA in each layer and that from the global BA in the top layer. For a $l$-th layer pyramid, the number of local windows in $i$-th layer ($i<l$) is $\frac{N}{s^i}$ and each local window consumes $O(w^3)$ of time. With $n$ number of parallel threads, the total time consumption of the local BA equals to the sum of the local BA in each layer which is $w^3\cdot\left(\sum^{l-1}_{i=1}\frac{N}{s^i}\cdot\frac{1}{n}\right)$ and the global BA in the $l$-th layer takes $O\left(\left(\frac{N}{s^{l-1}}\right)^3\right)$ of time. In summary, $O\left(T_l\right)$ is expressed as
\begin{equation}\label{eq:time_complexity}
  T_l=
  \left\{
  \begin{aligned}
  &N^3 &(l=1) \\
  &\frac{w^3}{n}\cdot\sum^{l-1}_{i=1}\frac{N}{s^i}+\left(\frac{N}{s^{l-1}}\right)^3 &(l>1)
  \end{aligned}
  \right.
\end{equation}
We view $T_l$ as a function of $l$ and calculate the optimal $l^*$ by letting the derivative of $T_l$ equal to zero, which leads to
\begin{equation}\label{eq:optimal_l}
  l^*=\left \lfloor\frac{1}{2}\log_s {\left(\frac{3N^2\left(s^3-s\right)n}{w^3}\right)}\right\rfloor
\end{equation}

\begin{figure}[ht]
  \centering
  \includegraphics[width=1.0\linewidth]{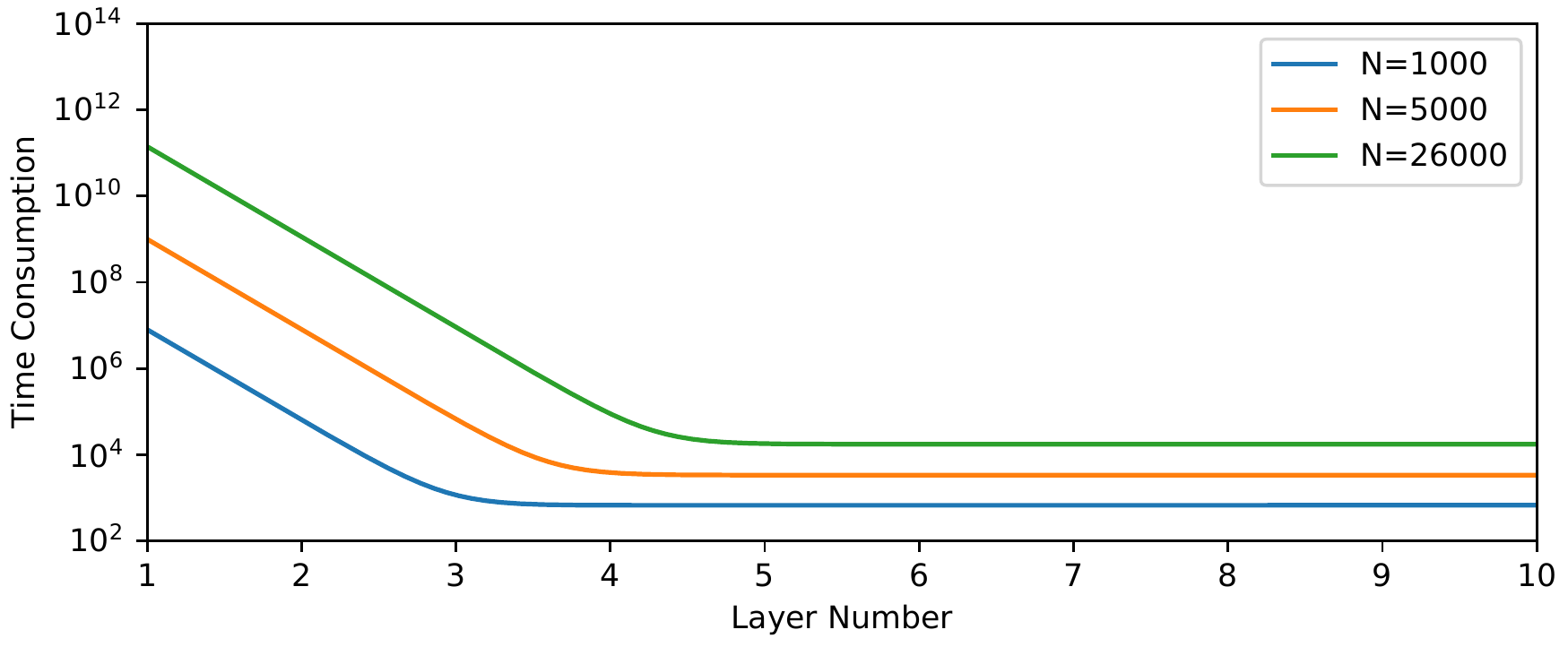}
  \caption{Example plot of time consumption $T_l$ w.r.t. the layer number $l$ and the pose number $N$ using \textit{w=10}, \textit{s=5} and \textit{n=8}. It is seen the overall time consumption is significantly reduced from the original BA ($l$=1) to our proposed hierarchical BA ($l$=3 or $l$=4).}
  \vspace{-0.4cm}
  \label{fig:Tl}
\end{figure}

Fig.~\ref{fig:Tl} shows an example of the computation time $T_l$ versus the layer number $l$ at different frame number $N$. As can be seen, as the layer number increases from $l$=1 (original BA) to $l^*$, the computation time is greatly reduced, suggesting the effectiveness of the proposed hierarchical BA. When $l > l^*$, the computation time does not increase much and keeps almost constant, suggesting that any layer number greater than $l^*$ will work equally well.


For feature extraction and association in each layer of the bottom-up hierarchical BA, we use an adaptive voxelization method proposed in~\cite{balm}, which extracts plane features of different sizes suitable for environments of different structures. To extract these various size plane features, the entire point cloud, after being transformed into the same global frame using the initial pose trajectory, is split into multiple voxels of size $V$, each goes through a plane test by checking the minimum and maximum eigenvalue ratio of the contained points is smaller than a threshold, i.e., $\frac{\lambda_1}{\lambda_3}<\theta$. If the plane test passes, points in the voxel will be viewed as lying on the same plane and used in the BA. Otherwise, the voxel will be recursively split until the contained points form a plane.

The above adaptive voxelization process is time-consuming when the number of points is extremely large. To mitigate this issue, we notice that points that are not considered as plane features in lower layers will not form a plane in upper layers either. We thus use only the plane feature points from each voxel in local BA to construct the keyframe for the upper layer in the bottom-up process. This procedure further saves time for next-layer adaptive-voxel map construction and improves the computation accuracy in local BA.

\vspace{-0.4cm}
\subsection{Top-Down Pose Graph Optimization}\label{sec:top_down}

The top-down pose graph optimization process aims to reduce the pose estimation errors in the bottom-up hierarchical BA process, which considers only features co-visible in the same local windows but ignores those observed across different local windows. As shown in Fig.~\ref{fig:lidar_frame}, the pose graph is constructed in a top-down manner in the pyramid structure. In each layer of the pyramid, the factors are relative poses between adjacent frames. Specifically, the cost item in the $i$-th layer factor between node $\mathbf{x}^i_j$ and $\mathbf{x}^i_{j+1}$ is defined as
\begin{equation}\label{eq:layer_factor}
\begin{aligned}
  \mathbf{e}^i_{j,j+1}&=\text{Log} \left(\left(\mathbf{T}^{i\ast}_{j,j+1}\right)^{-1}\left(\mathbf{T}^i_j\right)^{-1}\mathbf{T}^i_{j+1}\right)^\vee\\
  c\left(\mathbf{x}^i_j,\mathbf{x}^i_{j+1}\right)&=\left(\mathbf{e}^i_{j,j+1}\right)^T\left(\mathbf{\Omega}^i_{j,j+1}\right)^{-1}\mathbf{e}^i_{j,j+1}
\end{aligned}
\end{equation}
\begin{figure}[ht]
  \centering
  \includegraphics[width=1.0\linewidth]{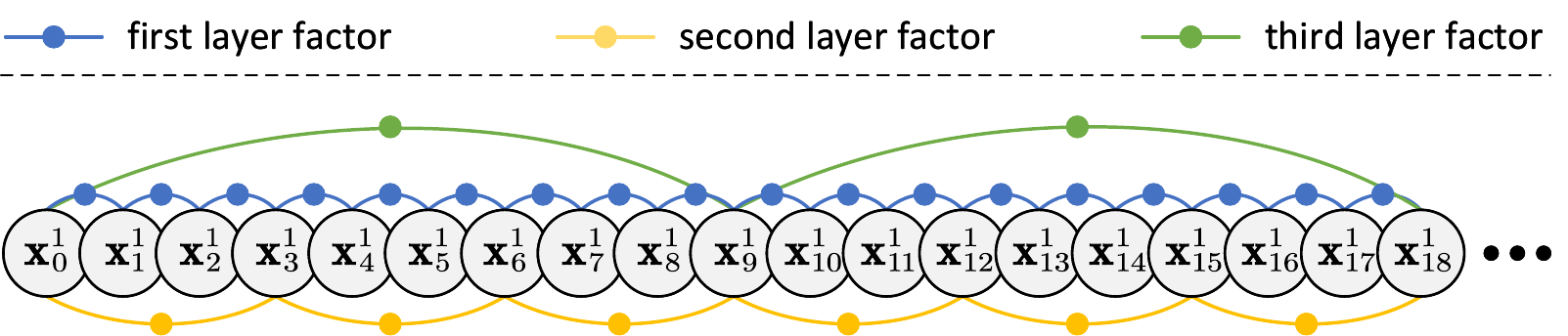}
  \caption{Final factor graph of our proposed approach with \textit{layer number l=3}, \textit{stride size s=3} and \textit{window size w=6}.}\label{fig:factor_graph}
  \vspace{-0.4cm}
  \label{fig:final_factor_graph}
\end{figure}
where $\mathbf{\Omega}^i_{j,j+1}$ is correlation matrix between pose $\mathbf{x}^i_j$ and $\mathbf{x}^i_{j+1}$, $i\in\mathcal{L}$ and $j\in\mathcal{F}^i$, and is computed by inverting the Hessian matrix $\mathbf{H}$ obtained from the bottom-up hierarchical BA process. The $\mathcal{L}=\{1,\cdots,l\}$ represents the set of $l$ layers and $\mathcal{F}^i=\{0,\cdots,N_i-1\}$ represents the set of $N_i$ numbers. Since the node $\mathbf x^{i+1}_j$ and $\mathbf x^i_{s\cdot j}$ are essentially the same, we have
\begin{equation*}
  \mathbf{T}^i_j=\mathbf{T}^{i-1}_{s\cdot j}=\cdots=\mathbf{T}^1_{s^{i-1}\cdot j},\ \forall i\in\mathcal{L}, j\in\mathcal{F}^i
\end{equation*}
Therefore, the cost item in~\eqref{eq:layer_factor} reduces to
\begin{equation}
\begin{aligned}
  &\mathbf{e}^i_{j,j+1}=\text{Log}\left(\left(\mathbf{T}^{i\ast}_{j,j+1}\right)^{-1}\left(\mathbf{T}^1_{s^{i-1}\cdot j}\right)^{-1}\mathbf{T}^1_{s^{i-1}\cdot(j+1)}\right)^\vee\\
  &c\left(\mathbf{x}^1_{s^{i-1}\cdot j},\mathbf{x}^1_{s^{i-1}\cdot(j+1)}\right)=\left(\mathbf{e}^i_{j,j+1}\right)^T\left(\mathbf{\Omega}^i_{j,j+1}\right)^{-1}\mathbf{e}^i_{j,j+1}
\end{aligned}
\end{equation}
where $i\in\mathcal{L}$ and $j\in\mathcal{F}^i$, and the original pose graph in Fig.~\ref{fig:lidar_frame} reduces to Fig.~\ref{fig:final_factor_graph}. It is noted that when $s<w$, the frames appearing within the overlap area of consecutive local windows could contribute multiple cost items (each local window contributes one). The objective function to be minimized is thus
\begin{equation}
  f\left(\mathbb{F},\mathcal{X}\right)= \sum_{i\in\mathcal{L}} \sum_{j\in\mathcal{F}^i} c\left(\mathbf{x}^1_{s^{i-1}\cdot j},\mathbf{x}^1_{s^{i-1}\cdot(j+1)}\right)
\end{equation}
where $\mathbb{F}=\{\mathbb{F}^1_i\mid i\in\mathcal{F}^1\}$ is the collection of all first layer frames. This factor graph is then solved by the Levenberg-Marquardt method using GTSAM\footnote{https://github.com/borglab/gtsam}.

\vspace{-0.2cm}
\section{Experiments}

\subsection{Accuracy Analysis}\label{sec:accuracy_analysis}

\subsubsection{Initial Odometry with Loop Closure}

In this section, we take the odometry results from the state-of-the-art (SOTA) LiDAR SLAM algorithms~\cite{mulls,fast_lio2,lio_sam} as the input and further optimize them with our proposed work. We demonstrate that our work could both improve the mapping quality (consistency) and the pose estimation accuracy even when the initial poses have already been pose-graph optimized. The BA and pyramid parameters used in all our following experiments, without further specification, are listed in Table~\ref{parameter}. The optimal $l^\ast$ is obtained by calculation using~\eqref{eq:optimal_l} based on the actual pose number $N$ in each sequence. For data length $N<5\times10^3$ (KITTI~\cite{kitti}, MulRan~\cite{mulran} and Newer College~\cite{ncd2021}), we have $l^*=3$ and for larger sequence, i.e., $N\geq5\times10^3$ (New College~\cite{ncd2020}), we have $l^*=4$.

\begin{table}[ht]
\centering
\caption{Hierarchical Bundle Adjustment Parameter Setting}
\begin{tabular}{ccc}
\toprule
Parameter & Value & Description \\
\midrule
$w$ & 10 & window size \\
$s$ & 5 & stride size \\
$n$ & 8 & threads number for parallel processing \\
$\theta_{local}$ & 0.05 & eigenvalue ratio threshold for local BA \\
$V_{local}$ & 4 & initial voxel size for local BA \\
$\theta_{global}$ & 0.1 & eigenvalue ratio threshold for global BA \\
$V_{global}$ & 4 & initial voxel size for global BA \\
\bottomrule
\end{tabular}
\label{parameter}
\vspace{-0.2cm}
\end{table}

\begin{table*}[ht]
\caption{RMSE of the ATE ($^\circ$/$m$) on KITTI Dataset with Loop Closure}
\centering{
\setlength{\tabcolsep}{0.5mm}{
\begin{tabular}{lcccccccccccc}
\toprule
Method & Seq. 00$^\ast$ & Seq. 01  & Seq. 02$^\ast$ & Seq. 03 & Seq. 04  & Seq. 05$^\ast$ & Seq. 06$^\ast$ & Seq. 07$^\ast$ & Seq. 08$^\ast$ & Seq. 09$^\ast$ & Seq. 10 & Avg. \\
\midrule
\textbf{Proposed} & \textbf{0.7}/\textbf{0.8} & \textbf{0.9}/\textbf{1.9} & \textbf{1.2}/5.1 & \textbf{0.7}/\textbf{0.6} & \textbf{0.1}/0.8  & \textbf{0.5}/\textbf{0.4} & \textbf{0.4}/\textbf{0.2} & \textbf{0.4}/\textbf{0.3} & 1.3/2.7 & \textbf{0.8}/1.3 & \textbf{0.6}/1.1 & \textbf{0.7}/\textbf{1.4}  \\
MULLS~\cite{mulls} & \textbf{0.7}/1.1 & \textbf{0.9}/\textbf{1.9} & \textbf{1.2}/5.4 & \textbf{0.7}/0.7 & 0.2/0.9  & 0.6/1.0 & \textbf{0.4}/0.3 & \textbf{0.4}/0.4 & 1.3/2.9 & 1.0/2.1 & \textbf{0.6}/1.1 & \textbf{0.7}/1.6  \\
LiTAMIN2~\cite{LiTAMIN2} & 0.8/1.3 & 3.5/15.9 & 1.3/\textbf{3.2} & 2.6/0.8 & 2.3/0.7  & 0.7/0.6 & 0.8/0.8 & 0.6/0.5 & \textbf{0.9}/\textbf{2.1} & 1.7/2.1 & 1.2/\textbf{1.0} & 1.2/2.4  \\
SuMa~\cite{SuMa} & \textbf{0.7}/1.0 & 3.2/13.8 & 1.7/7.1 & 1.5/0.9 & 1.8/\textbf{0.4}  & 0.5/0.6 & 0.7/0.6 & 1.1/1.0 & 1.2/3.4 & 0.8/\textbf{1.1} & 0.9/1.3 & 1.1/3.2 \\
\bottomrule
\end{tabular}}}
\begin{tablenotes}
  \item \qquad\qquad\quad$^\ast$ Sequence with loops.
\end{tablenotes}
\label{kitti_w_lc}
\vspace{-0.3cm}
\end{table*}

\begin{figure}[ht]
  \centering
  \includegraphics[width=1.0\linewidth]{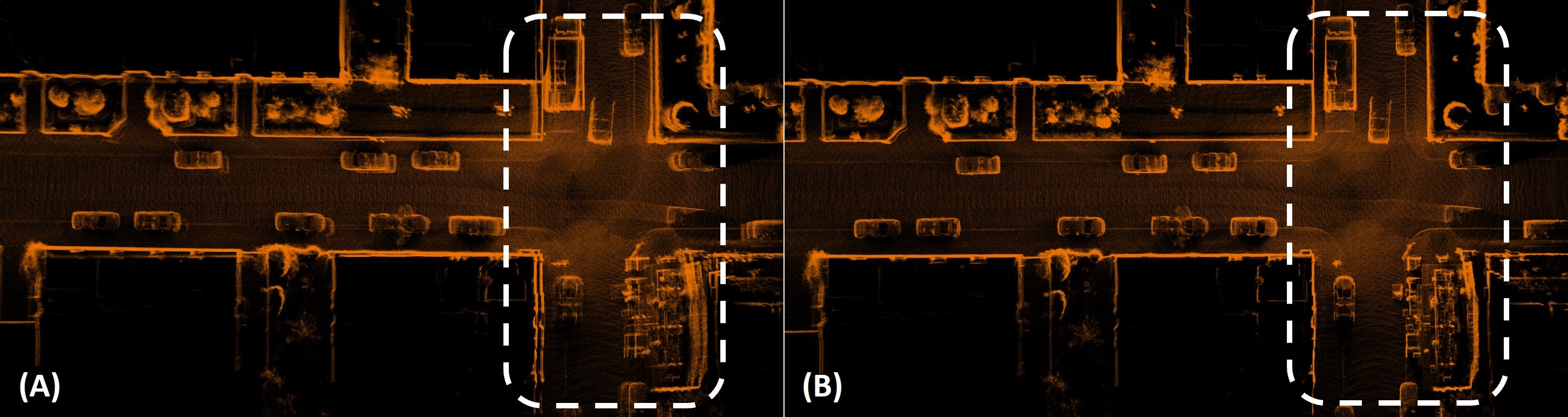}
  \caption{Mapping result from (A) MULLS~\cite{mulls} and (B) our proposed method on KITTI Seq. 07. The white dashed rectangle emphasizes the divergence.}
  \vspace{-0.4cm}
  \label{fig:kitti07}
\end{figure}

\begin{figure}[ht]
  \centering
  \includegraphics[width=1.0\linewidth]{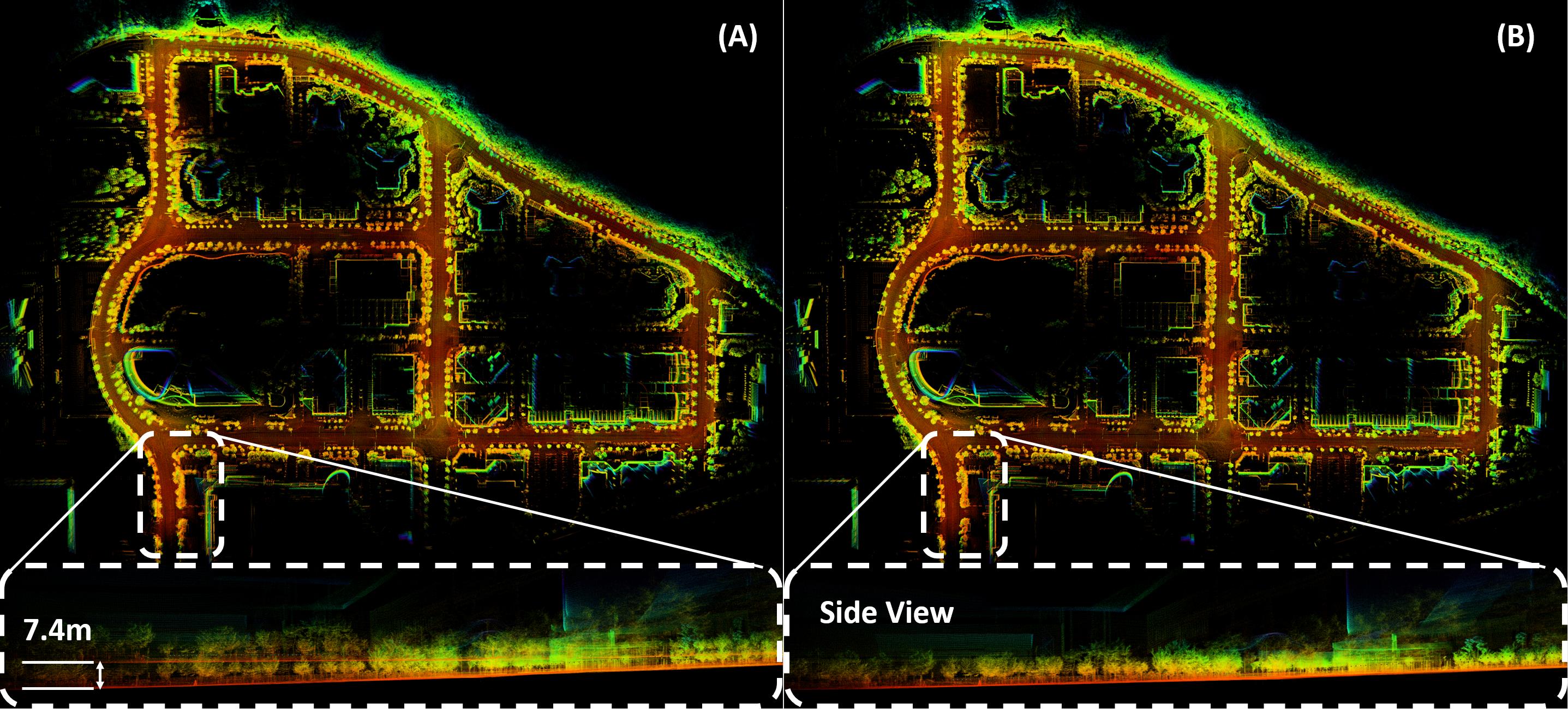}
  \caption{Point cloud mapping in sequence DCC03 of MulRan. (A): The mapping result from LIO-SAM~\cite{lio_sam} with loop closure. (B): The mapping result from our proposed method. The start (red) and ending (blue) positions have been emphasized by the white dashed rectangle and enlarged at the bottom (zoomed view is recommended).}
  \label{fig:dcc03}
\end{figure}

We first test on the KITTI dataset~\cite{kitti} and use the pose estimation results from MULLS with loop closure~\cite{mulls} as our initial input. The RMSE of the absolute rotation (degree) and translation (meter) errors are summarized in Table~\ref{kitti_w_lc}. We choose the absolute trajectory error (ATE) as the evaluation criterion since each LiDAR frame has one unique ground truth. As can be seen, our proposed work could further improve the pose estimation accuracy, especially in translation, even when they are pose-graph optimized. Though our approach does not achieve the best result in every sequence, our work produces the optimal ATE results (0.7$^\circ$/1.4m) on average compared with other SOTA methods. Moreover, since the pose graph optimization does not directly optimize the consistency of the point cloud, the divergence in the map is not fully eliminated in~\cite{mulls} (see Fig.~\ref{fig:kitti07}). This divergence could be iteratively solved with our proposed hierarchical BA and pose graph optimization, which in return, reduces the ATE on pose estimations.

\begin{table}[ht]
\caption{\centering RMSE of the ATE ($m$) on MulRan Dataset with Loop Closure}
\centering{
\setlength{\tabcolsep}{1mm}{
\begin{tabular}{lrrrr}
\toprule
\multicolumn{2}{l}{Sequence} & \textbf{Proposed} & LIO-SAM~\cite{lio_sam} & LEGO-LOAM~\cite{lego_loam} \\
\midrule
\multirow{3}{*}{DCC} & 01 & \textbf{5.20} & 5.67 & 6.95 \\
 & 02 & \textbf{3.22} & 3.48 & 5.49 \\
 & 03 & \textbf{2.54} & 2.84 & 6.29 \\
\midrule
\multirow{3}{*}{KAIST} & 01 & \textbf{3.36} & 3.55 & 5.45 \\
 & 02 & \textbf{3.75} & 3.81 & 5.49 \\
 & 03 & \textbf{3.53} & 3.60 & 5.70 \\
\midrule
\multirow{3}{*}{\begin{tabular}[c]{@{}r@{}}RIVER\\-SIDE\end{tabular}} & 01 & \textbf{8.92} & 9.25 & 19.05 \\
 & 02 & \textbf{7.94} & 8.04 & 16.04 \\
 & 03 & \textbf{10.26} & 10.37 & 30.91 \\
\midrule
\multicolumn{2}{l}{Avg.} & \textbf{5.41} & 5.62 & 11.62 \\
\bottomrule
\end{tabular}}}
\label{MulRan_w_lc}
\end{table}

We then test our proposed method on another spatially large-scale spinning LiDAR dataset, MulRan~\cite{mulran}. The average lengths of the contained sequences DCC, KAIST and RIVERSIDE, are 4.9km, 6.1km and 6.8km, respectively. Unlike the KITTI dataset, which collects most of the data within the urban area, MulRan dataset includes more challenging scenes from the viaduct, river and woods. We choose to use the pose-graph optimized pose trajectory results from LIO-SAM~\cite{lio_sam} as our input. The RMSE of the absolute translation error has been summarized in Table~\ref{MulRan_w_lc}. It is seen our proposed work could still improve the pose estimation accuracy regardless of these challenging unstructured woods scenes. However, these scenes make the LIO-SAM generate poor relative pose and covariance estimations, which further leads to partial failure in the loop closure, causing a large divergence in altitude (see Fig.~\ref{fig:dcc03}). With our proposed hierarchical BA and pose graph optimization mechanism, this divergence could be fully eliminated, and the pose trajectory accuracy is thus further improved.


Lastly, we test our work on the sequence $\textit{long\_experiment}$ which is the longest sequence ($N$=26557) in the New College dataset~\cite{ncd2020}. We choose the FAST-LIO2~\cite{fast_lio2} to provide the pose trajectory as our initial input. It is noted that these initial poses are generated without loop closure. Table~\ref{ncd_lc} shows the absolute translation error of our proposed and other SOTA methods. It is seen our proposed work outperforms other loop closure enabled SOTA methods and achieves the optimal accuracy on this large time-scale sequence.

\begin{table}[ht]
\caption{\centering RMSE of the ATE ($m$) on Newer College Dataset with Loop Closure}
\centering{
\setlength{\tabcolsep}{1mm}{
\begin{tabular}{lccc}
\toprule
Method & long\_experiment \\
\midrule
\textbf{Proposed} & \textbf{0.26} \\
GICP Matching Factor~\cite{gpu_accelerated} & 0.28 \\
FAST-LIO2~\cite{fast_lio2} & 0.35 \\
LIO-SAM~\cite{lio_sam} & 0.53 \\
CT-ICP~\cite{ct_icp} & 0.58 \\
\bottomrule
\end{tabular}}}
\label{ncd_lc}
\vspace{-0.3cm}
\end{table}

\subsubsection{Initial Odometry without Loop Closure}

\begin{table*}[ht]
\caption{RMSE of the ATE ($^\circ$/$m$) on KITTI Dataset without Loop Closure}
\centering{
\setlength{\tabcolsep}{0.5mm}{
\begin{tabular}{lcccccccccccc}
\toprule
Method & Seq. 00$^\ast$ & Seq. 01  & Seq. 02$^\ast$ & Seq. 03 & Seq. 04  & Seq. 05$^\ast$ & Seq. 06$^\ast$ & Seq. 07$^\ast$ & Seq. 08$^\ast$ & Seq. 09$^\ast$ & Seq. 10 & Avg. \\
\midrule
\textbf{Proposed} & 1.0/\textbf{1.2} & \textbf{1.0}/\textbf{2.4}  & 2.3/9.0  & \textbf{0.7}/\textbf{0.6} & \textbf{0.2}/0.9  & \textbf{0.5}/\textbf{0.7} & \textbf{0.3}/\textbf{0.2} & \textbf{0.4}/\textbf{0.3} & 1.3/2.5 & \textbf{0.7}/\textbf{1.5} & \textbf{0.5}/1.1 & \textbf{0.8}/\textbf{1.9} \\
MULLS~\cite{mulls} & 1.7/6.1 & \textbf{1.0}/\textbf{2.4}  & 2.4/10.7 & \textbf{0.7}/0.7 & \textbf{0.2}/0.9  & 1.0/2.4 & 0.4/0.6 & 0.5/0.6 & 1.9/4.3 & 1.0/3.1 & \textbf{0.5}/1.1 & 1.0/3.0 \\
Voxel Map~\cite{voxel_map} & \textbf{0.9}/2.8 & 1.9/7.8 & \textbf{1.7}/\textbf{6.1} & 1.2/0.7 & 0.6/\textbf{0.3} & 0.8/1.2 & 0.4/0.4 & 0.7/0.7 & \textbf{1.1}/\textbf{2.3} & 1.0/1.9 & 1.0/1.1 & 1.2/2.9 \\
SuMa~\cite{SuMa} & 1.0/2.9 & 3.2/13.8 & 2.2/8.4  & 1.5/0.9 & 1.8/0.4  & 0.7/1.2 & 0.4/0.4 & 0.7/0.5 & 1.5/2.8 & 1.1/2.9 & 0.8/1.3 & 1.4/3.9 \\
LiTAMIN2~\cite{LiTAMIN2} & 1.6/5.8 & 3.5/15.9 & 2.7/10.7 & 2.6/0.8 & 2.3/0.7  & 1.1/2.4 & 1.1/0.9 & 1.0/0.6 & 1.3/2.5 & 1.7/2.1 & 1.2/\textbf{1.0} & 1.8/5.1 \\
\bottomrule
\end{tabular}}}
\begin{tablenotes}
  \item \qquad\qquad\quad$^\ast$ Sequence with loops.
\end{tablenotes}
\label{kitti_wo_lc}
\vspace{-0.3cm}
\end{table*}

\begin{figure}[ht]
  \centering
  \includegraphics[width=1.0\linewidth]{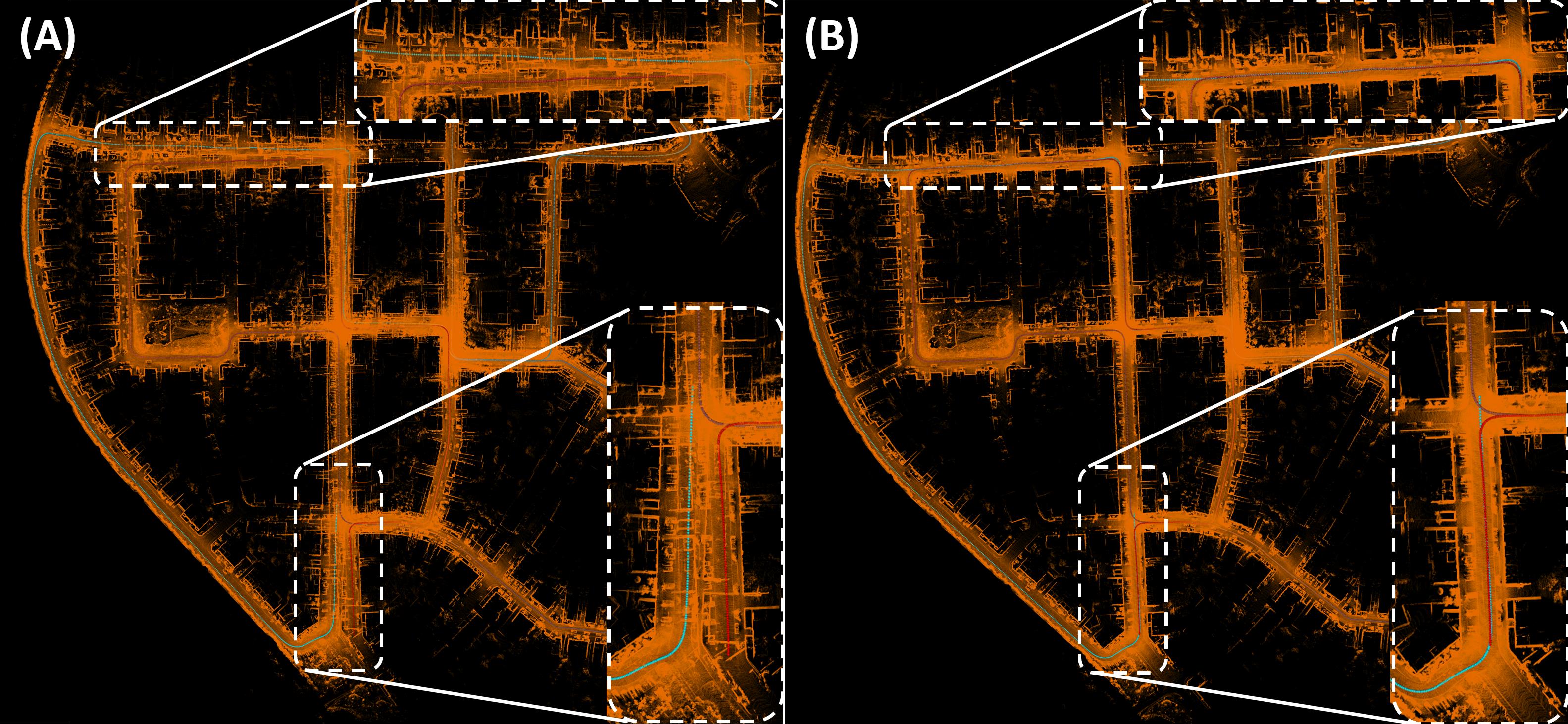}
  \caption{Closure of the gap on KITTI dataset Seq. 00 with our proposed method. The mapping result (A) is provided by MULLS~\cite{mulls} without loop closure, and (B) our proposed method. The odometry is colored by the moving distance from the start (red) to the end (blue). The main gaps are detailed by white dashed rectangles. The full experiment video is available on https://youtu.be/CuLnTnXVujw.}
  \vspace{-0.2cm}
  \label{fig:kitti00}
\end{figure}

\begin{figure}[ht]
  \centering
  \includegraphics[width=1.0\linewidth]{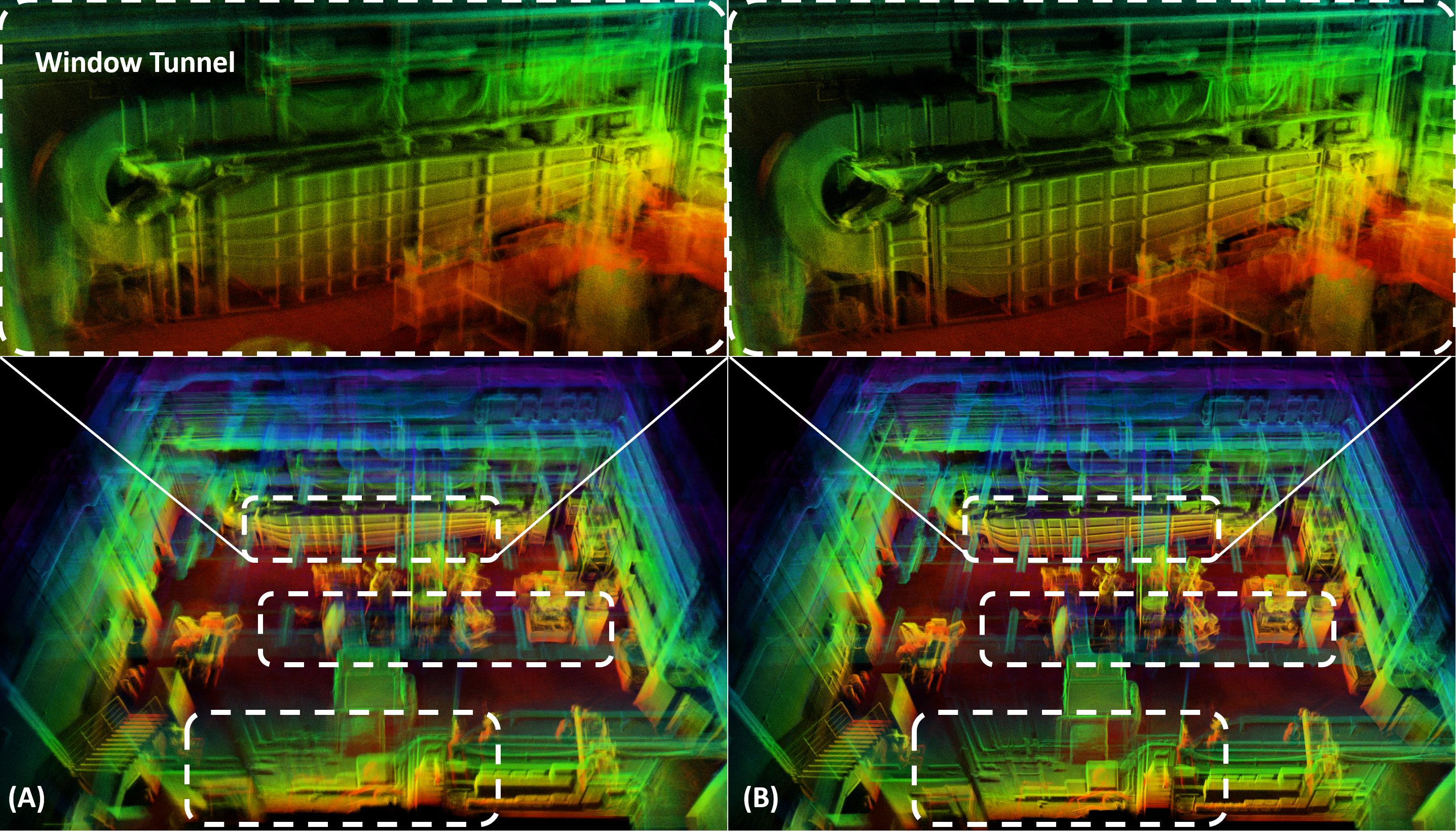}
  \caption{Reconstructed point cloud map of scene-1 using (A) FAST-LIO2~\cite{fast_lio2} (MME=-2.99) and (B) our proposed work (MME=-3.06). The main differences are emphasized by the white dashed rectangles, including the walls (bottom), ceiling lights (middle) and wind tunnel (top). Please view our experiment video for more details.}
  \vspace{-0.2cm}
  \label{fig:lg_machine}
\end{figure}

In this section, we demonstrate that our proposed work can converge well even when the loop is not closed in the initial pose trajectory. We first test on the KITTI dataset~\cite{kitti} using the initial pose trajectory estimated from MULLS~\cite{mulls} without loop closure. The RMSE of the absolute rotation and translation errors are summarized in Table~\ref{kitti_wo_lc}. As can be seen, other SOTA methods get much worse ATEs due to the lack of loop closure function, whereas our proposed work could still produce reliable pose estimations, with the absolute rotation and translation errors being largely reduced (e.g., Seq. 00 and Seq. 08) and achieving the optimal ATE on average (0.8$^\circ$/1.9m). This is due to the reason that, in local BA, the strict parameters (see Table~\ref{parameter}) ensure the frames within each local window do not diverge, and the loose parameters of the top layer global BA could implicitly identify potential divergences, generating correct relative pose constraints (see Fig.~\ref{fig:kitti00}). Though false feature correspondence matching in global BA might happen that incorrect top layer factor is added to the pose graph, the dense bottom layer factors ensure this incorrect factor will not drag the poses away from the correct direction.

\begin{figure}[ht]
  \centering
  \includegraphics[width=1.0\linewidth]{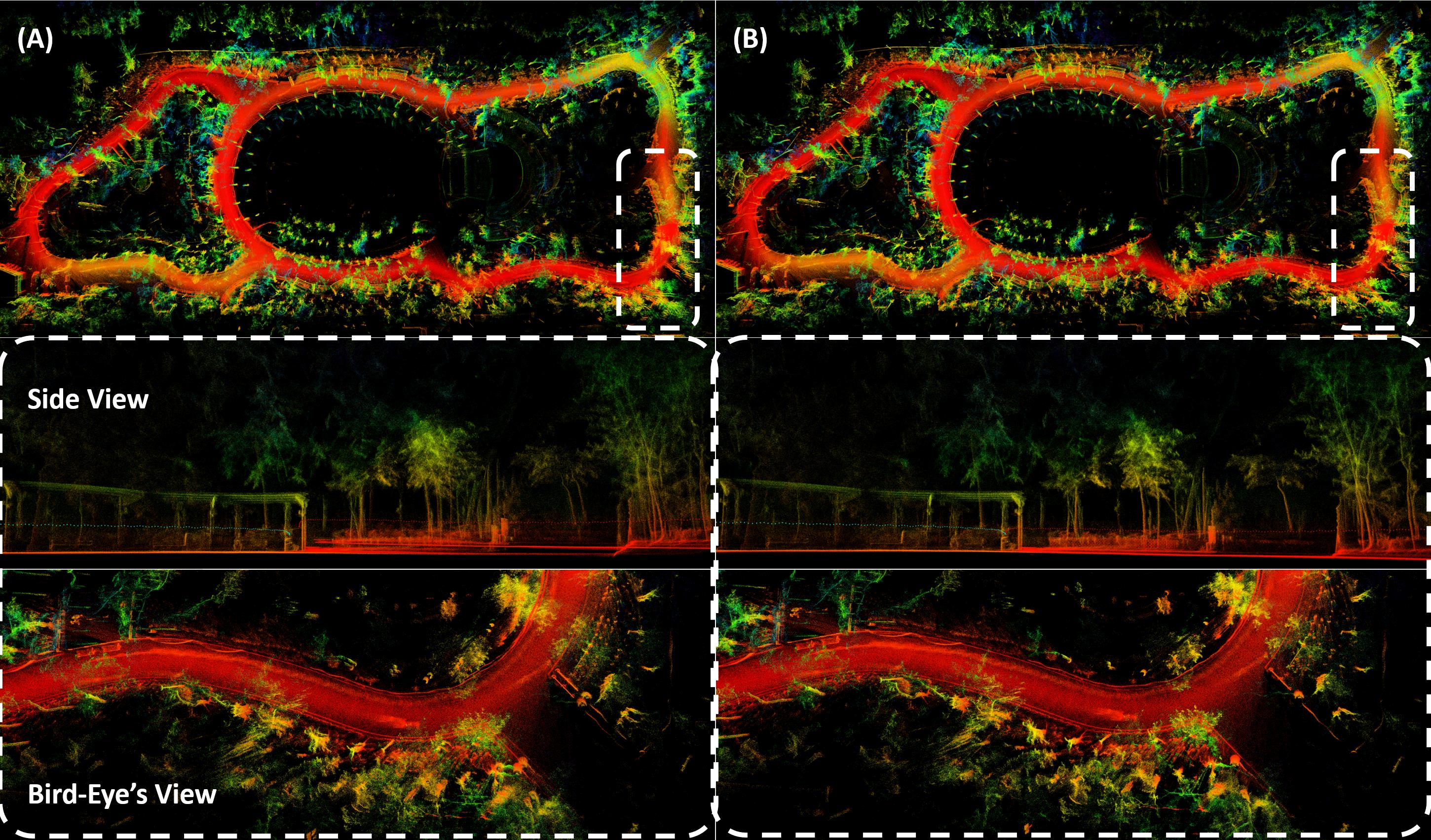}
  \caption{Reconstructed point cloud map of scene-2 using (A) FAST-LIO2~\cite{fast_lio2} (MME=-2.60) and (B) our proposed work (MME=-2.69). The main differences are emphasized by the white dashed rectangle. The second row depicts the side view of divergence in height. The third row shows the divergence from the bird-eye's view.}
  \vspace{-0.4cm}
  \label{fig:park}
\end{figure}

We further validate the versatility of our proposed work on our self-collected dataset using solid-state LiDAR~\cite{lowcost} in both structured and unstructured scenes. The first test scene is a structured indoor factory with several irregular-shaped pipelines and machines (see Fig.~\ref{fig:lg_machine}). The size of this scene is around 14$\times$16$\times$8m, and the length of this sequence is 7339 frames. The pyramid parameters are set the same as in Table~\ref{parameter} except that the initial voxel size is decreased due to the smaller size of the indoor scene ($V_{local}$=$V_{global}$=1m). The second test scene is an unstructured outdoor park with bush, grassland and woods around (see Fig.~\ref{fig:park}). The size of this scene is around 95$\times$195m, and the length of this sequence is 3407 frames. We thus adjust the initial voxel size ($V_{local}$=$V_{global}$=2m) without changing the other pyramid parameters from Table~\ref{parameter}. For both test sequences, we use the pose estimations from FAST-LIO2~\cite{fast_lio2} without loop closure as our input. Since the measurement of the ground truth is not available, we instead use the mean map entropy (MME~\cite{MME}) to evaluate the mapping quality (see Table~\ref{mme}). Since the MME calculates the natural logarithm of the determinant of the covariance matrix, the smaller MME is, the more consistent the point cloud is. It is seen our proposed work could further improve the mapping consistency and close the gap in both structured and unstructured scenes regardless of the LiDAR types.

\begin{table}[ht]
\caption{MME on Self Collected Dataset}
\centering
\setlength{\tabcolsep}{1mm}{
\begin{tabular}{lcc}
\toprule
Method & scene-1 & scene-2 \\
\midrule
FAST-LIO2~\cite{fast_lio2} & -2.99 & -2.60 \\
\textbf{Proposed} & \textbf{-3.06} & \textbf{-2.69} \\
\bottomrule
\vspace{-0.6cm}
\end{tabular}}
\label{mme}
\end{table}

\vspace{-0.5cm}
\subsection{Ablation Study}

\subsubsection{Pose Graph Optimization versus Direct Assign}

\begin{table*}[ht]
\caption{RMSE of the ATE ($^\circ/m$) on KITTI Dataset}
\centering
\setlength{\tabcolsep}{1mm}{
\begin{tabular}{lcccccccccccc}
\toprule
Method & Seq. 00$^\ast$ & Seq. 01 & Seq. 02$^\ast$ & Seq. 03 & Seq. 04 & Seq. 05$^\ast$ & Seq. 06$^\ast$ & Seq. 07$^\ast$ & Seq. 08$^\ast$ & Seq. 09$^\ast$ & Seq. 10 & Avg. \\
\midrule
Direct Assign$^\star$ & 0.68/0.81 & \textbf{0.87}/\textbf{1.88} & 1.44/5.25 & 0.72/\textbf{0.62} & 0.19/\textbf{0.81} & \textbf{0.51}/0.52 & \textbf{0.35}/0.29 & \textbf{0.40}/\textbf{0.26} & 1.28/2.76 & 0.86/1.59 & \textbf{0.55}/1.13 & 0.71/1.47 \\
\textbf{Proposed}$^\star$ & \textbf{0.67}/\textbf{0.79} & \textbf{0.87}/1.91 & \textbf{1.19}/\textbf{5.08} & \textbf{0.71}/0.63 & \textbf{0.14}/0.82 & 0.53/\textbf{0.39} & 0.36/\textbf{0.22} & \textbf{0.40}/0.30 & \textbf{1.25}/\textbf{2.68} & \textbf{0.83}/\textbf{1.26} & 0.57/\textbf{1.12} & \textbf{0.68}/\textbf{1.38} \\
\midrule
Direct Assign & 1.30/1.32 & \textbf{0.97}/\textbf{2.37} & 2.18/9.47 & 0.67/\textbf{0.58} & 0.28/\textbf{0.87} & 0.66/0.83 & 0.35/0.31 & 0.47/0.32 & 1.71/3.42 & 1.02/2.12 & 0.54/1.08 & 0.92/2.06 \\
\textbf{Proposed} & \textbf{1.00}/\textbf{1.17} & 0.98/2.41 & 2.26/8.95 & \textbf{0.65}/0.59 & \textbf{0.21}/0.90 & \textbf{0.48}/\textbf{0.74} & \textbf{0.33}/\textbf{0.22} & \textbf{0.43}/\textbf{0.29} & \textbf{1.29}/\textbf{2.53} & \textbf{0.70}/\textbf{1.47} & \textbf{0.53}/\textbf{1.07} & \textbf{0.81}/\textbf{1.85} \\
\bottomrule
\end{tabular}}
\label{ablation_kitti}
\begin{tablenotes}
  \item \quad $^\ast$ Sequence with loops.
  \item \quad $^\star$ The initial poses trajectory is generated with loop closure.
\end{tablenotes}
\end{table*}

\begin{figure}[ht]
  \centering
  \includegraphics[width=1.0\linewidth]{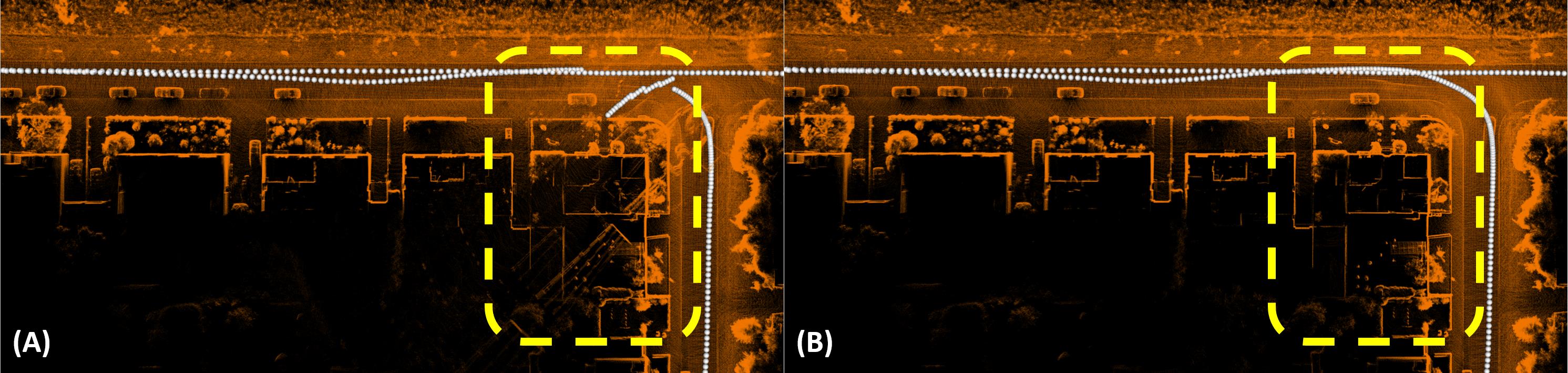}
  \caption{Mapping results (A) without and (B) with our proposed pose graph optimization in KITTI Seq. 08. The main difference is emphasized by the yellow dashed rectangle. The white dots represent the trajectory. Inconsistency occurs at the higher layer and is iteratively assigned to the bottom layer poses if no pose graph optimization is applied.}
  \vspace{-0.5cm}
  \label{fig:kitti08}
\end{figure}

In this section, we demonstrate our proposed top-down design is non-trivial. To update the bottom layer poses, a trivial method is to directly assign the optimized upper layer poses to the lower layer poses, e.g., an optimized pose of a keyframe from the upper layer is used to update the first $s$ poses from the lower layer within the corresponding local window. For example, in Fig.~\ref{fig:lidar_frame}, the poses $\mathbf{x}^2_0,\mathbf{x}^2_1,\mathbf{x}^2_2$ are updated by $\mathbf{x}^{3\ast}_0$ and the poses $\mathbf{x}^2_3,\mathbf{x}^2_4,\mathbf{x}^2_5$ are updated by $\mathbf{x}^{3\ast}_1$, respectively. We test this trivial method (``Direct Assign") with our proposed pose graph optimization mechanism on KITTI dataset~\cite{kitti} both using the poses generated before and after loop closure from MULLS~\cite{mulls} as the input. The RMSE of the absolute rotation and translation errors are summarized in Table~\ref{ablation_kitti}. As can be seen, our proposed pose graph optimization outperforms the direct assigning one both in pose estimation accuracy and mapping consistency (see Fig.~\ref{fig:kitti08}). Though the above trivial strategy could still improve the pose estimation accuracy, such a method neglects the relative pose constraints from each local window, e.g., the pose $\mathbf{x}^2_3$ is involved in two local windows, whereas it is updated by $\mathbf{x}^{3\ast}_1$ only without considering the relative constraint from $\mathbf{x}^2_2$. This will lead to a mapping inconsistency between frames $\mathbb{F}^2_2$ and $\mathbb{F}^2_3$ and further between frames $\mathbb{F}^1_8$ and $\mathbb{F}^1_9$. In our proposed pose graph optimization approach, the first layer factor ensures the consistency between every adjacent frame while the second and above layer factors ensure the gap is converged towards the correct direction.

\subsubsection{Hierarchical BA versus Reduced BA}

In this section, we demonstrate that our proposed bottom-up design is non-trivial. To accelerate the Hessian matrix solving process, a trivial way is to keep only the block diagonal elements (of size $s$ of the stride length) of the original Hessian matrix and solve this reduced matrix without considering the relative pose constraints among different local windows as in our method. We verify this reduced BA with the original BA and our approach on the DCC sequence of the MulRan dataset~\cite{mulran}. The RMSE of the absolute translation error and the total optimization time of all methods are summarized in Table~\ref{ablation_dcc}. It is seen our proposed work achieves the optimal precision as the original BA method while drastically reducing the computation time. This is due to the reason that the original BA needs to construct an adaptive voxel map using all points (adaptive-voxel map), which quickly escalates as the involved pose number increases. Our method conducts BA in local windows, so we only have to construct an adaptive voxel map using a very small amount of points in the local window, and different local windows can be paralleled. The reduced BA actually takes longer time than the original BA due to two reasons. First, it needs to construct an adaptive-voxel map similar to the original BA. Second, the reduced BA zeros the off-diagonal block elements, which leads to inaccurate Hessian estimation and significantly slows down the convergence speed. In our experiments, the reduced BA may even fail to converge when the maximum iteration number is reached ($\textit{max\_iter=10}$) while the original BA converges within a few steps. Moreover, since we use relatively strict parameters on the local BA, factors from these layers ensure that glitches will not appear between every adjacent frame. For the simplified and the original BA, only the strict parameter could be adopted, otherwise, false feature matching will frequently happen (points within the voxel do not form a plane feature).

\begin{table}[ht]
\caption{\centering RMSE of ATE ($m$) and Optimization Time on DCC Sequence of MulRan Dataset}
\centering
\setlength{\tabcolsep}{1mm}{
\begin{tabular}{lrrrrrr}
\toprule
\multirow{2}{*}{Method} & \multicolumn{2}{c}{DCC01} & \multicolumn{2}{c}{DCC02} & \multicolumn{2}{c}{DCC03} \\
 & RMSE & Time & RMSE & Time & RMSE & Time \\
\midrule
Initial & 5.67m & - & 3.48m & - & 2.84m & - \\
Original BA & \textbf{5.20m} & 2897.54s & \textbf{3.20m} & 2853.76s & \textbf{2.54m} & 7219.38s \\
Reduced BA & 5.66m & 4972.10s & 3.46m & 4923.90s & 2.83m & 8176.46s \\
\textbf{Proposed} & \textbf{5.20m} & \textbf{206.04s} & 3.22m & \textbf{328.59s} & \textbf{2.54m} & \textbf{256.27s} \\
\bottomrule
\end{tabular}}
\label{ablation_dcc}
\end{table}

\vspace{-0.4cm}
\subsection{Computation Cost}\label{sec:computation_cost}

\begin{figure}[ht]
  \centering
  \includegraphics[width=1.0\linewidth]{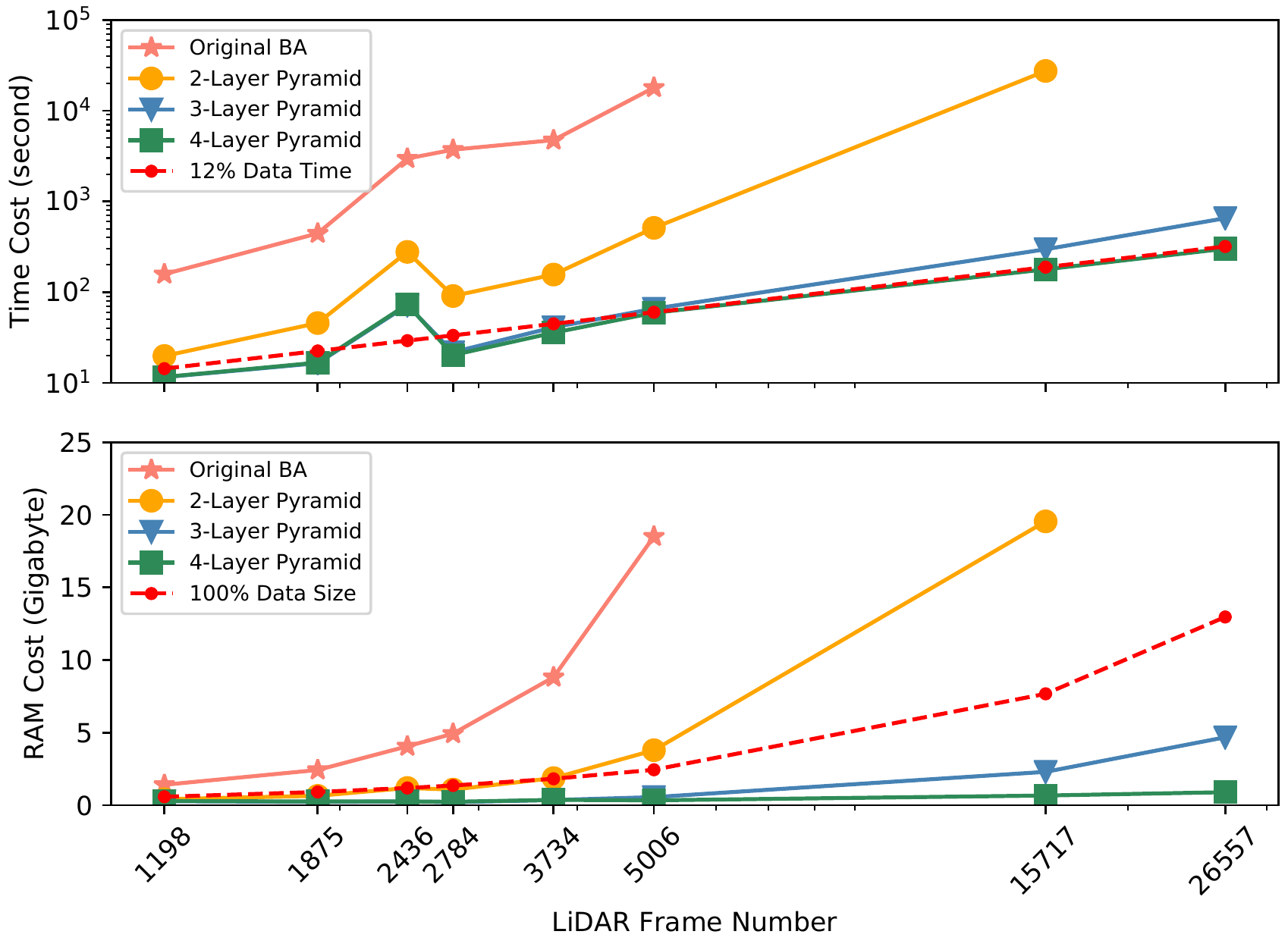}
  \caption{Comparison of computation time and RAM costs with multiple setups of layers used in the pyramid under Newer College Dataset~\cite{ncd2020,ncd2021}. Different setups are represented using different colors. The red dashed lines represent 12$\%$ of the total data time and total point cloud size of the corresponding sequence.}
  \vspace{-0.4cm}
  \label{fig:computation_time_ram}
\end{figure}

In this section, we demonstrate that our proposed approach is computationally efficient, especially on the large-scale dataset. We test our proposed method with multiple setups of used layers on New College~\cite{ncd2020} and Newer College~\cite{ncd2021} datasets whose data length varies from $10^3$ to $10^4$ frames. The total computation time (including the adaptive-voxel map construction and BA time) and the maximum RAM memory consumption recorded for each setup at all sequences are illustrated in Fig.~\ref{fig:computation_time_ram}. Due to the huge time and RAM consumption of the original BA method, we do not test it on the last two sequences since the plot could already depict the trend.

It is seen for all test scenes the more layers are used, the less computation time the pyramid takes. When the pose number $N<5\times10^3$, the time and RAM consumption of 3-layer and 4-layer pyramids are similar, and when $N>5\times10^3$, the 4-layer pyramid becomes optimal. All these phenomenons are in accordance with our theoretical analysis shown in Fig.~\ref{fig:Tl} and \eqref{eq:optimal_l}. Since the third test scene ($N$=2436) contains a more complex environment (thus more adaptive-voxel map construction time), the total time consumption is actually larger than the latter scene. Despite this, by choosing the optimal layer setup, our work could converge within around $12\%$ of the whole data time and consumes much smaller RAM during operation, which is suitable for practical usage.

\vspace{-0.4cm}
\section{Conclusion}

In this paper, we propose a hierarchical BA and pose graph optimization-based work to optimize the pose estimation accuracy and mapping consistency globally for the large-scale LiDAR point cloud. With the bottom-up hierarchical BA, we parallelly solve multiple BA problems with a much smaller Hessian matrix size than the original BA method. With the top-down pose graph optimization, we smoothly and efficiently update the LiDAR poses without generating glitches. We validate the effectiveness of our work on spatially and timely large-scale LiDAR datasets with structured and unstructured scenes, given a good initial pose trajectory or with large drifts. We demonstrate our proposed work outperforms other SOTA methods in pose estimation accuracy and mapping consistency on multiple public spinning LiDAR and our self-collected solid-state LiDAR datasets. In our future work, we could combine the IMU pre-integration and LiDAR measurement noise model into our hierarchical BA work.

\bibliographystyle{unsrt}
\bibliography{reference}

\end{document}